\documentclass[runningheads]{llncs}
\usepackage{graphicx}
\usepackage{color, soul} 
\usepackage{wrapfig}
\usepackage{enumitem}
\usepackage{times}
\usepackage{epsfig}
\usepackage{amsmath}
\usepackage{amssymb}
\usepackage{multirow}
\usepackage[table]{xcolor}
\newcolumntype{C}[1]{>{\centering\arraybackslash}m{#1}}
\definecolor{almond}{rgb}{0.94, 0.87, 0.8}
\graphicspath{{"./Figures/"}} 

\begin{document}
\title{Trajectory Prediction by Coupling Scene-LSTM with Human Movement LSTM}
%
%
\author{Manh Huynh and Gita Alaghband}
%

%
\institute{Department of Computer Science and Engineering\\ 
Univesity of Colorado Denver\\
\email{\{manh.huynh,gita.alaghband\}@ucdenver.edu}}
\maketitle              
\begin{abstract}
We develop a novel human trajectory prediction system
that incorporates the scene information (Scene-LSTM) as
well as individual pedestrian movement (Pedestrian-LSTM)
trained simultaneously within static crowded scenes. We
superimpose a two-level grid structure (grid cells and subgrids)
on the scene to encode spatial granularity plus common
human movements. The Scene-LSTM captures the
commonly traveled paths that can be used to significantly
influence the accuracy of human trajectory prediction in local
areas (i.e. grid cells). We further design scene data
filters, consisting of a hard filter and a soft filter, to select
the relevant scene information in a local region when necessary
and combine it with Pedestrian-LSTM for forecasting
a pedestrian’s future locations. The experimental results
on several publicly available datasets demonstrate that our
method outperforms related works and can produce more
accurate predicted trajectories in different scene contexts.

\end{abstract}
\section{Introduction}
\label{sec:introduction}

\begin{wrapfigure}{r}{5cm}
	\centering
	\includegraphics[trim=0pt 12pt 0pt 0pt, width=\linewidth]{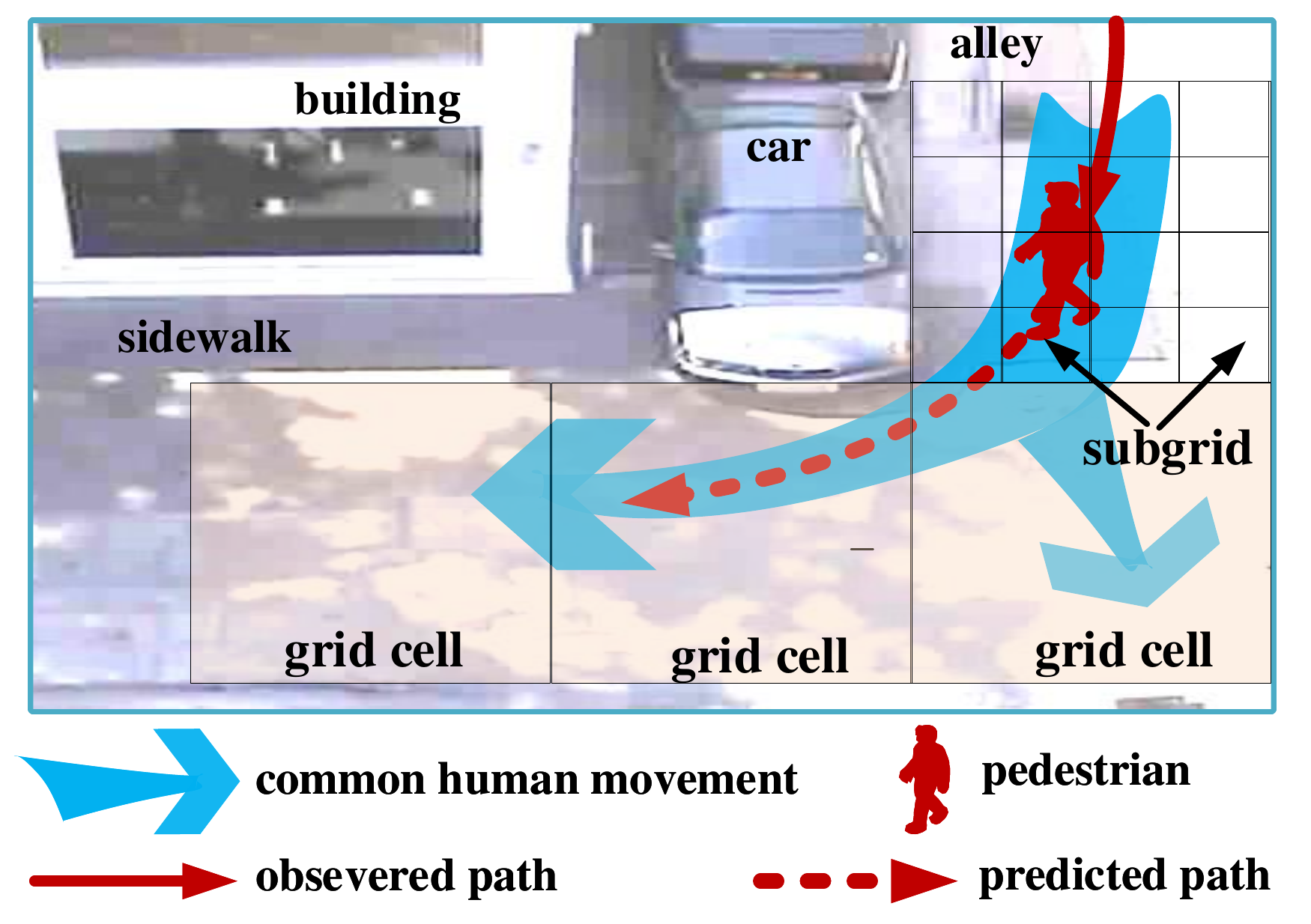}
	\caption{Scene-LSTM learns common human movements on a two-level grid structure. The common human movement is filtered and used in combination with individual movement (Pedestrian-LSTM) to predict a pedestrian’s future locations.}
	\label{fig:introduction}
\end{wrapfigure}

Human movement trajectory prediction is an essential task in computer vision with applications in autonomous driving cars \cite{levinson2011towards}, robotic navigation systems \cite{pellegrini2009you,vivacqua2017self}, and intelligent human tracking systems~\cite{leonard1990application,manh2018spatiotemporal}. Given the past movement trajectories of pedestrians in a video sequence, the goal is to predict their near future trajectories (lists of continuous two-dimensional locations) (Figure \ref{fig:introduction}). For the most part, predicting future human trajectories is challenging due to: \textbf{(i)} existence of many possible future trajectories, especially in open areas where people move and change directions freely at any time (multi-modal problem); \textbf{(ii)} social interactions (e.g. grouping, avoiding, etc.) can impact decisions of the next movements; \textbf{(iii)} structures within scenes can impose certain paths.

To deal with these challenges, several social-interaction methods~\cite{pellegrini2009you,alahi2016social,trautman2010unfreezing,vemula2018social,gupta2018social,robicquet2016learning,xu2018encoding} have been proposed. The traditional methods \cite{pellegrini2009you,trautman2010unfreezing,yamaguchi2011you} use hand-crafted features to characterize the social interactions.  Recently, several social-interaction methods \cite{alahi2016social,vemula2018social,gupta2018social,xu2018encoding} leverage the power of LSTM (Long Short-Term Memory) networks for modeling the individual movement behaviors and social interactions. Although social interaction has been shown to be effective in predicting future human locations in some scenarios, it does not perform well in multi-modal environments. For example, to avoid collisions with other people while walking, one can choose to go left or right. 

To partially handle this multi-modal problem, the contexts of a scene can be used.  Several proposed methods  \cite{xue2018ss,ballan2016knowledge,varshneya2017human,bartoli2018context} have gained improvements by extracting the scene’s visual features from video images and use them in combination with social-interaction features. They hypothesize that people may move in the same directions under similar scene contexts. The main limitation is that the low-level visual features cannot fully explain the human movements. Given the same scene layouts, there is still a high probability that people will choose different paths. Furthermore, generating similar visual features for similar scene contexts is difficult due to camera positions, angles, etc. Thus, these visual features are often used in combination with the social-interaction features to achieve the desired accuracy. 

In this paper, we propose and develop a novel scene model, called Scene-LSTM, to learn common human movement features in each grid cell (and at finer subgrid level when necessary) as shown in Figure \ref{fig:introduction}, which can be used in combination with individual movement (Pedestrian-LSTM) to predict a pedestrian’s future locations. A Scene Data Filter (SDF) is further designed to select the relevant scene information to predict the pedestrians’ next locations, based on their current locations and walking behaviors. The key components of the SDF are a “hard filter” and a “soft filter.” The hard filter makes decisions on whether the scene information should be used in predicting pedestrians’ future trajectories based on their current locations on the two-level grid structures (grid cells and subgrids). The filtered scene information from the hard filter is used by the soft filter for further processing. The soft filter selects the relevant scene information for each pedestrian, based on their movement behaviors, to predict future locations.

In summary, the contributions of this paper are threefold: \textbf{(1)} A new Scene-LSTM model is learned simultaneously with a LSTM-based human walking model; \textbf{(2)} a SDF selects relevant scene information to predict pedestrians’ trajectories with the help of hard filer and soft filter; \textbf{(3)} Evaluations on public datasets show that Scene-LSTM outperforms several related methods in terms of human trajectory prediction accuracy. Ablation studies are conducted to show the relevance and impact of each system component.


\section{Related Works}
\label{sec:related_works}

Research in predicting future human trajectories has been focused on modeling human-human interactions \cite{pellegrini2009you,alahi2016social,trautman2010unfreezing,vemula2018social,gupta2018social,robicquet2016learning,xu2018encoding}. There have been very few studies related to human-scene interactions \cite{xue2018ss,ballan2016knowledge,varshneya2017human,bartoli2018context}.  
   
\textbf{Human-human methods.} To model human-to-human interactions, some researchers \cite{pellegrini2009you,trautman2010unfreezing,yamaguchi2011you} characterize their social interactions as features and calculate the next locations of each pedestrian by minimizing some function of these features. For example, S.Pellegrini et al. \cite{pellegrini2009you} calculates the desired velocities of each pedestrian by minimizing the energy function of collision avoidance, speed, and direction towards the pedestrians’ final destinations. K. Yamaguchi et al. \cite{yamaguchi2011you} broadens the model \cite{pellegrini2009you} with social group behaviors such as attractions and groupings using energy functions which are minimized using gradient descent \cite{kingma2014adam}. P. Trautman et al. \cite{trautman2010unfreezing} characterizes human movements and collision potentials using Gaussian processes with multiple particles and apply maximum a posteriori (MAP) to minimize the collision potentials to yield the best next locations.  Although utilizing these social-interaction features helps predict future human movements, they are built upon specific social-interaction rules; thus, they do not apply well to all possible scenarios. 

Recently, several LSTM based methods \cite{alahi2016social,vemula2018social,gupta2018social,xu2018encoding} have been proposed to learn individual human movement behaviors and social interactions by leveraging the memorizing power of LSTM. For example, Social-LSTM \cite{alahi2016social} uses a social pooling layer to learn the social interactions of the main target and nearby pedestrians. Other methods \cite{vemula2018social,gupta2018social,xu2018encoding} model the social interactions in the entire scene, where people far-away from the main target may also have social impacts on this target’s movements. These methods use different types of network architectures such as structural recurrent neural network \cite{vemula2018social},  generative neural network \cite{goodfellow2014generative}, and deep neural network  \cite{xu2018encoding}.

\textbf{Human-scene methods.} A relatively small body of recent work have studied the impact of scene structures (e.g. buildings, static obstacles, etc.) on human trajectory prediction. These methods \cite{xue2018ss,ballan2016knowledge,varshneya2017human,bartoli2018context} combine scene features with social interactions to predict human movement trajectories. Some methods \cite{xue2018ss,varshneya2017human} extract feature of the scene layouts using Convolutional Neural Network. Ballan et al. \cite{ballan2016knowledge} utilizes several techniques (e.g. color histograms, scale-invariant feature transform (SIFT), etc.) to calculate scene visual descriptors in local (patch) and global (image) context. F. Bartoli et al. \cite{bartoli2018context} measures the distances between the targets and obstacles in the scene and combines them with Social-LSTM \cite{alahi2016social}.  These human-scene methods have made gains in improving prediction accuracy. However, the limitation is that the low-level scene visual features cannot fully capture the high-level scene contexts (e.g. common human movements), which can significantly improve the accuracy of human trajectory prediction as we  will present in this paper.

\section{System Design}
\label{sec:system_design}

\textbf{Problem Definition:} The problem under consideration is prediction of human movement trajectories in static crowded scenes. Let’s define $X_i^t=(x_i^t,y_i^t)$ as the spatial location of target i at time t, and N as the number of pedestrians in the number of observed frames, $T_{obs}$. The problem is stated as: given the trajectories of all pedestrians in observed frames ${(x_i^t,y_i^t )}$, where $t =1,…,T_{obs}$ and $i=1,…,N,$ predict the next locations for each pedestrian in the predicted frames $T_{pred}$.  [Time t corresponds to frame number.]

\begin{figure*}[t]
	\centering
	\includegraphics[width=1\linewidth]{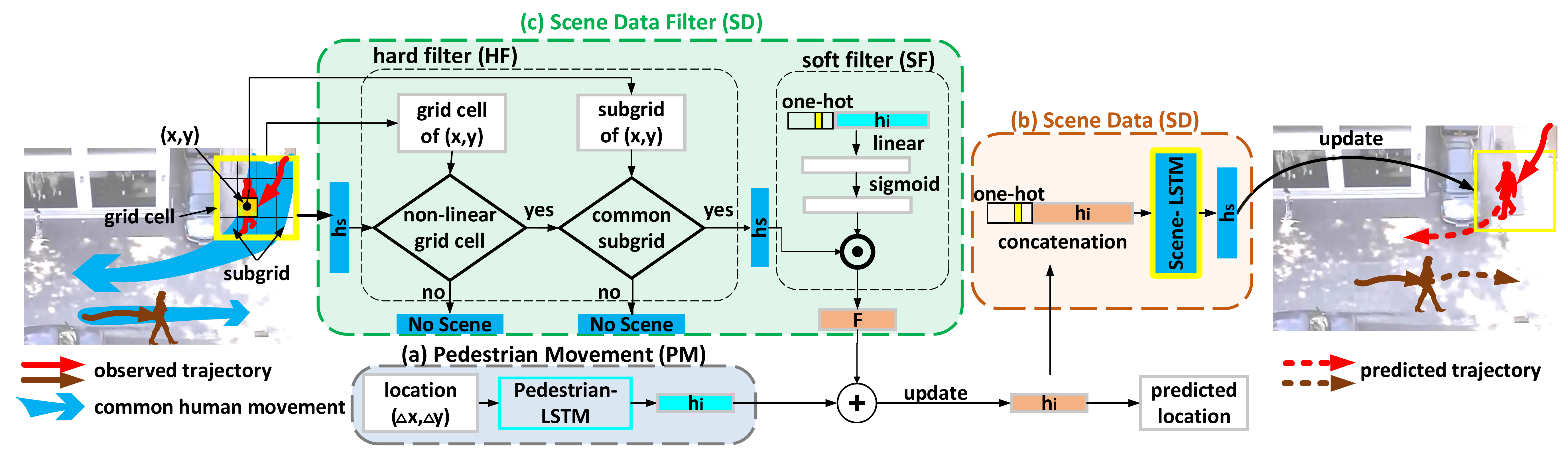}
	
	\caption{The system consists of three main modules: Pedestrian Movement (PM), Scene Data (SD) and Scene Data Filter
		(SDF). PM models the individual movement of pedestrians. SD encodes common human movements in each grid cell. SDF
		selects relevant scene data to update the Pedestrian-LSTM, which is used to predict the future locations. $\otimes$ denotes elementwise multiplication. $\oplus$ denotes vector addition. $h_i$ and $h_s$ are the hidden states of Pedestrian-LSTM and Scene-LSTM, respectively.}
	\label{fig:system_overview}
	\vspace{-5mm}
\end{figure*}

Our system design, depicted in Figure \ref{fig:system_overview}, consists of three main modules: Pedestrian Movement (PM), Scene Data (SD), and Scene Data Filter. The description of each module is explained below:
\label{sec:related_works}

\noindent \textbf{(a)} Pedestrian Movements (PM) module models the individual pedestrian’s movement behavior using a LSTM (Pedestrian-LSTM) (one LSTM/pedestrian). Pedestrian-LSTM utilizes its memory cell to remember the past movements of a pedestrian. For better adaptability across scenes, pedestrian’s relative locations with respect to the previous locations are used as inputs to the Pedestrian-LSTM network at each training step. 

\begin{figure}[t]
	\centering
		\includegraphics[width=\linewidth]{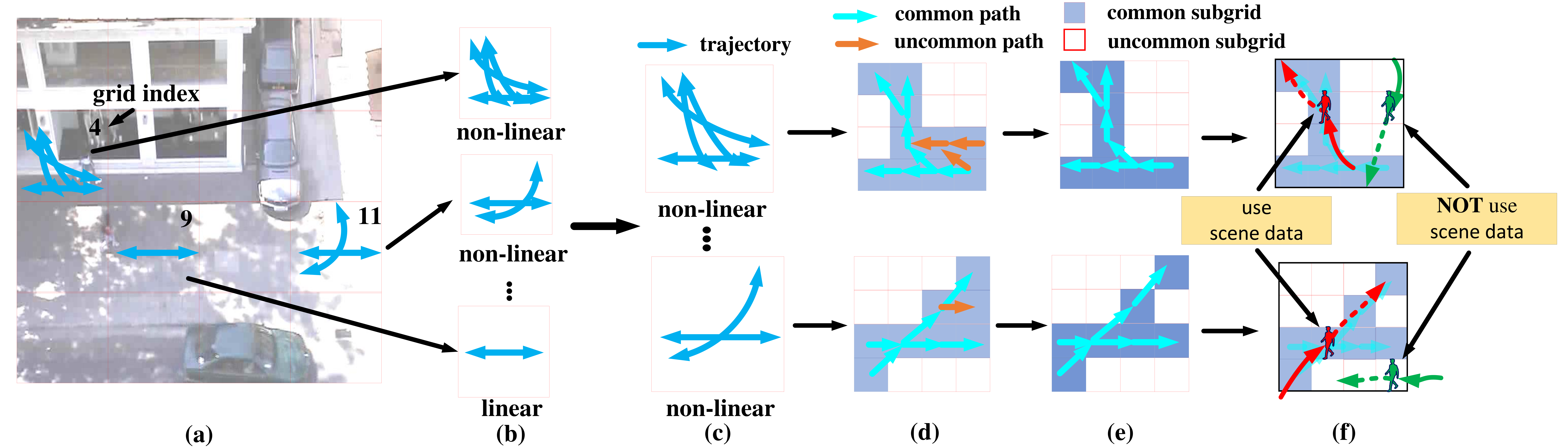}
	\caption{Illustrations of the hard filter, which determines whether the scene data should be applied in predicting the future locations of a pedestrian. (a) the frame image is first divided into $n \times n$ grid cells ($n = 4$ in this example) to capture all human movements in each grid cell; (b) \& (c) only non-linear grid cells are selected for further processing at the subgrid level; the scene data is not applied for pedestrians in the linear grid cell; (d) a non-linear grid cell is further divided into $m \times m$ subgrids ($m = 4$) and each trajectory is parsed into subgrid paths; (e) the common subgrids, occupied by common subgrid paths; (f) at prediction time, the decision of use/not use scene data depends on the current location of each pedestrian. If the pedestrian’s current location is in the common subgrids, the scene data is used (\textcolor{red}{red} pedestrian); otherwise, it is not used (\textcolor{green}{green} pedestrian).}
	\label{fig:hard_filter}
	\vspace{-5mm}
\end{figure}

\noindent \textbf{(b)} Although the PM is responsible for modeling the individual pedestrian’s movement behavior, there will be scenarios where the pedestrian model alone does not have adequate information to predict trajectories. In such cases, data from scene can help steer the prediction trajectories the right way. The Scene Data (SD) module models all human movements within the entire scene identifying commonly travelled paths at various movement granularities. The scene is superimposed with a two-level grid structure: grid cells which are further divided into subgrids.  The SD uses a LSTM (Scene-LSTM) (one/grid cell) to encode the pedestrians’ movements in each grid cell.  The absolute location $(x, y)$ of each pedestrian is used to locate them in the scene at the subgrid level. This is represented in the form of a one-hot vector described in the next section. The combination (i.e. concatenation) of the one-hot vector and the hidden state h of Pedestrian-LSTM is used as an input to Scene-LSTM. Although Scene-LSTMs are able to encode all human movements in each grid cell of a scene during the training process, we must recognize that using the combined scene and pedestrian data will not work well for all cases. The following two scenarios describe when the scene data should not be used to influence next prediction: \textbf{(1)} The scene data is not needed for predicting the future locations of the pedestrians whose movements are linear and therefore not impacted by the scene structures. For example, the pedestrians in open areas (e.g. grid cell 9, Figure \ref{fig:hard_filter}a) mostly walk linearly without any scene structure constraints. The scene information has no effect on the pedestrians’ movements in these areas. \textbf{(2)} The scene information in the grid cells where various past trajectories coincide may be unhelpful in predicting the human future locations. This is because the memories of these grid cells, encoding all different types of trajectories, do not learn any specific common movements and worsen the prediction accuracy.

\noindent \textbf{(c)} To handle the aforementioned challenges, we propose the Scene Data Filter (SDF) which consists of two filters: a hard filter (HF) and a soft filter (SF). The HF helps us decide whether the scene data of a grid cell should be applied to predict a given pedestrian’s next location based on this pedestrian’s current grid cell and subgrid locations. This is done based on whether the grid cell is labeled as linear or non-linear during training/observation period; Figure \ref{fig:hard_filter}b shows.

A grid cell is characterized as “linear” if all human trajectories in this grid cell are linear. In other words, people always make linear movements in this cell. Prediction of future trajectory of a pedestrian that travels in a linear grid cell is simple and can rely on Pedestrian-LSTM only. However, pedestrians travelling in a non-linear grid cell which contains non-linear trajectories, may have varying paths caused by social interactions and/or scene structures. This scenario can be captured by the subgrid structure of the non-linear grid cells. The HF in this case is used to enforce the coupling of the scene data with the pedestrian model to predict a pedestrian’s next location based on this person’s current subgrid location and common path.

The intuition of the HF at the subgrid level granularity is based on the observation that if pedestrians walk in the common subgrid paths, there is a high probability that they tend to follow the same path.  In this case, the scene data will be used in conjunction with the pedestrian data.  A common subgrid path, as shown in Figure \ref{fig:hard_filter}d, is a path between two subgrids commonly travelled by a number of pedestrians greater than a pre-defined threshold $p=3$. The subgrids travelled by common paths are called common subgrids (Figure \ref{fig:hard_filter}e). If the pedestrian’s current location is in a common subgrid, the scene information will be applied; otherwise, only the PM is used (Figure \ref{fig:hard_filter}f). It is important to note that once the common subgrids are selected, the HF not only will capture the common movements caused by the scene constraints or social interactions, but also implicitly excludes all uncommon movements that can degrade the prediction. The selections of non-linear/linear grid cells and common paths is done by processing the training data only once at the pre-processing step, while the hidden states h and memory cells c in Pedestrians-LSTMs and Scene-LSTMs are updated at every training step.

\begin{figure}[t]
	\centering
		\includegraphics[width=\linewidth]{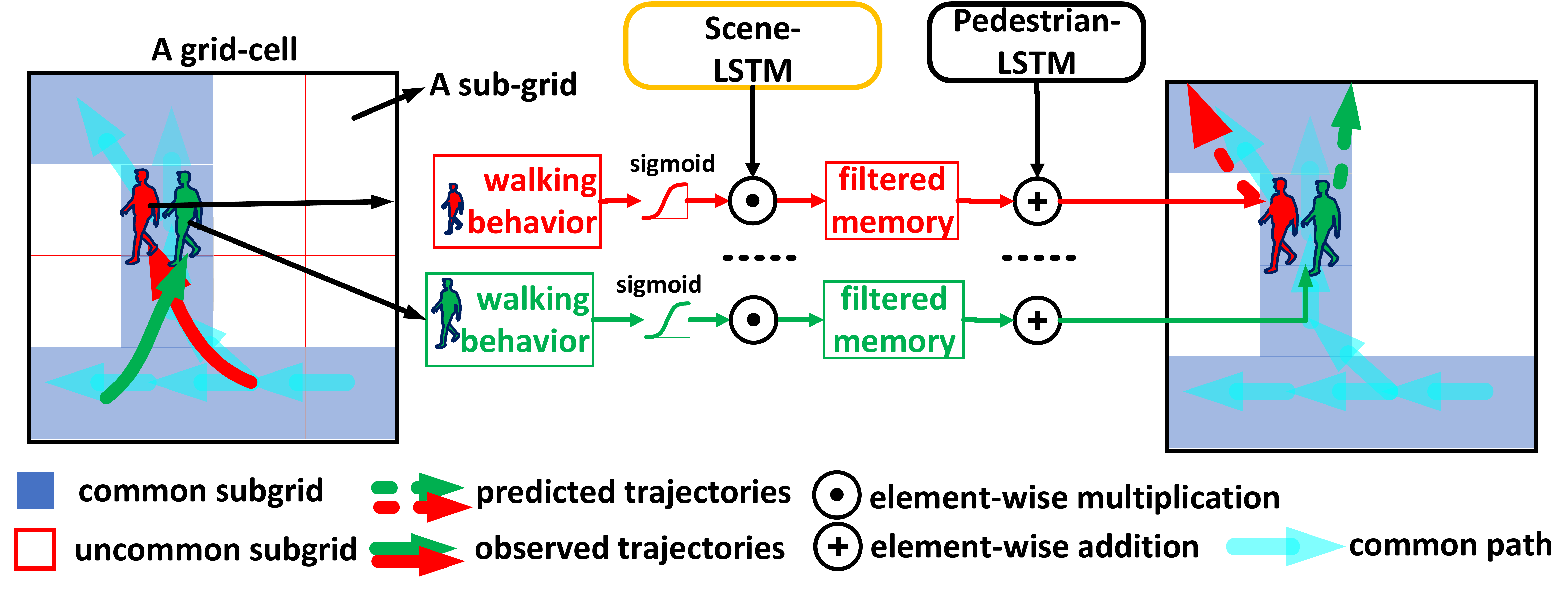}
	\caption{ Illustrations of the soft filter. The relevant information of scene data (i.e. Scene-LSTM) is selected using each pedestrians’ walking behavior. The filtered grid-cell memory of each pedestrian is then used in combination with pedestrian movements (Pedestrian-LSTM) to predict the future trajectories. }
	\label{fig:soft_filter}
	\vspace{-5mm}
\end{figure}

Lastly, the soft filter (SF) processes the scene data (obtained from HF) for each pedestrian based on their movement behaviors. As shown in Figure \ref{fig:soft_filter}, although two pedestrians (red and green) step on the same common subgrids at the same time, they travel in different paths in the future. Thus, the relevant scene data should be selected for each pedestrian depending on their past movements. The relevant scene data is then used to update the hidden state h of Pedestrian-LSTM and predict the next locations. The updated hidden state h of a pedestrian is also used to update the scene data of the non-linear grid cell where this person walks in. 
\section{Implementation Details}
\label{sec:implementation}
\noindent \textbf{Pedestrian Movement Module (PM)}. In this paper, we adopt a LSTM network similar to the one proposed in Social-LSTM \cite{alahi2016social} to model each pedestrian’s movement behavior. Given the relative location $(\Delta{x}_i^t,\Delta{y}_i^t) = (x_i^t,y_i^t )-(x_i^{t-1},y_i^{t-1})$ of person $i$ at time $t$, we embed it to get a fixed length vector $e_i^t$ and use it as an input to learn this person’s LSTM state $(h_i^t,c_i^t)$: 
\begin{align}
e_i^t &= \phi(W_{ie},[\Delta{x}_i^t,\Delta{y}_i^t]) \\
(h_i^t,c_i^t) &= \text{LSTM}((h_i^{t-1},c_i^{t-1}),e_i^t;W_{pm})
\end{align}
where $\phi(\cdot)$ is embedding function with ReLU non-linearity. $W_{ie}$ denotes embedding weights. $W_{pm}$ denotes LSTM weights and are shared among all pedestrians. 

\noindent \textbf{Scene Data Module (SD).} The Scene-LSTM $(h_g^t,c_g^t)$ of grid cell $g$ at time $t$ is updated as: 
\vspace{-1mm}
\begin{gather}
		V_i^t = O(x_i^t,y_i^t ) \\
		(h_g^t,c_g^t) = \text{LSTM}((h_g^{t-1},c_g^{t-1}),[V_i^t,h_i^t]; W_g )
\end{gather}
where $W_g$ denotes LSTM weight matrices, $O(\cdot)$ is a function to convert the absolute location $(x_i^t,y_i^t)$ of pedestrian $i$  to a one-hot vector $V_i^t$. The one-hot vector $V_i^t$ represents the relative location of this person corresponding to a subgrid within the grid cell $g$. In order to calculate $V_i^t$, each grid cell is further divided into $m \times m$ subgrids; thus, $V_i^t$ has size $m \times m$ with values $[0,…1,…0]$, where $1$ indicates the subgrid that this target occupies. The concatenation of $V_i^t$ and $h_i^t$, $([V_i^t,h_i^t])$, represents the current walking behavior and location of pedestrian $i$ in grid cell $g$. Thus, the grid cell’s memory, encoding this information, captures all human movements, which can be used to predict the human future locations.

\noindent \textbf{Hard filter.} The illustrations of the hard filter (HF) at the grid cell and the subgrid level for various scenarios are shown in Figure \ref{fig:hard_filter}. The trajectories in each grid cell of a scene are collected at pre-processing step for training data (Figure \ref{fig:hard_filter}a). Each grid cell is classified as “linear” or “non-linear” (Figure \ref{fig:hard_filter}b). The linear grid cell indicates that the scene data will not be applied to predict human future locations in this grid cell while the non-linear grid cells are selected (Figure \ref{fig:hard_filter}c) to be processed at the subgrid level. At the subgrid level, each trajectory in a non-linear grid cell is parsed into subgrid paths, which are then classified as common and uncommon subgrid paths (Figure \ref{fig:hard_filter}d). The common subgrid paths define the common subgrids (Figure \ref{fig:hard_filter}e). At prediction time, if a pedestrian steps in the common subgrids (e.g. \textcolor{red}{red} pedestrian, Figure \ref{fig:hard_filter}f), the scene data will be applied to predict this person’s next location; otherwise, no scene data will be used (\textcolor{green}{green} pedestrian).

\noindent \textbf{Soft filter (SF).} The final filtered scene data $F_i^t$ for pedestrian $i$ at time $t$ is calculated as: 
\begin{gather}
S_i= \sigma(linear([V_i^t,h_i^t])) \\
F_i^t=S_i  \odot h_g^t
\end{gather}
where $h_g^t$ and $h_i^t$ are the hidden states of the Scene-LSTM of the non-linear grid cell and Pedestrian-LSTM, respectively. $S_i$, the soft-filter vector of pedestrian $i$, is calculated by first concatenating one-hot vector $V_i^t$ and $h_i^t$. It is further processed using a linear layer, followed by a sigmoid function to convert $S_i $ within range $[0,1]$. The final filtered scene data $F_i^t$ is a result of element-wise multiplication ($\odot$) between $S_i$ and the scene data from hard filter $h_s^t$. 

Finally, $F_i^t$  is used to update hidden state $h_i^t$ (obtained from the Pedestrian-LSTM) of pedestrian $i$. $h_i^t$ is then used to predict the next location of this person: 
\begin{gather}
h_i^t= h_i^t+F_i^t \\
(\mu_i^{t+1}, \sigma_i^{t+1}, p_i^{t+1}) = W_{of}h_i^t\\
(\Delta \hat{x}_i^{t+1}, \Delta \hat{y}_i^{t+1} ) \sim \aleph(\mu_i^{t+1}, \sigma_i^{t+1}, p_i^{t+1}) \\
(\hat{x}_i^{t+1}, \hat{y}_i^{t+1}) = (\hat{x}_i^{t} + \Delta \hat{x}_i^{t+1}, \hat{y}_i^{t} + \Delta \hat{y}_i^{t+1})
\end{gather}
where $W_{of}$ is a weight matrix. Similar as \cite{alahi2016social}, the bivariate Gaussian distribution  $\aleph(\mu_i^{t+1}, \\ \sigma_i^{t+1}, p_i^{t+1})$
 is used to predict the next locations. model is trained by minimizing the negative log-likelihood loss $L$ \cite{kingma2014adam}:
\begin{gather}
L(W) = - \Sigma_{i=0}^N\Sigma_{t=0}^T \log(P(x_i^t,y_i^t |\mu_i^t, \sigma_i^t, p_i^t))
\end{gather}
where $W$ is the set of weight matrices. $N$ is number of targets, $T=T_{obs}+T_{pred}$ is the number of frames used for training. $(x_i^t,y_i^t)$ is the true location of target $i$  at time $t$. By minimizing $L(W)$, the likelihood that the predicted location $(\hat{x}_i^{t}, \hat{y}_i^{t}) $ is closer to the true location $(x_i^t,y_i^t)$ is maximized.

\section{Evaluation}
\label{sec:evaluation}

\noindent \textbf{Datasets:} As with the related prior research \cite{alahi2016social,gupta2018social,xue2018ss,vemula2018social}, we first evaluate our model on two publicly available datasets: ETH \cite{lerner2007crowds} and UCY \cite{pellegrini2009you}. These datasets contain 5 video sequences (ETH-Hotel, ETH-Univ, UCY-Univ, ZARA-01, and ZARA-02) consisting of 1536 pedestrians in total with different movement patterns and social interactions: people crossing each other, avoiding collisions, or moving in groups. These sequences are recorded in 25 frames/second (fps) and contain 4 different scene backgrounds. To present the generalizability of our method on other datasets, we further evaluate on: Town Center \cite{benfold2011stable} (1 video) and PETS09S2 \cite{ferryman2009pets2009} (3 videos), and Grand Central \cite{zhou2012understanding} (1 video).  The Town Center and the PETS09S2 datasets consist of short-duration videos, originally used for human tracking, and consist of considerable amounts of social interactions (e.g. collision avoidings  and group walkings). We also test our model on a long-duration video (33:20 mins) of  Grand Central dataset, which consists of densely social interactions and was originally  used for crowd behavior analysis.

\noindent \textbf{Metrics:} we evaluate our system using three metrics, introduced by S. Pellegrini et al. \cite{pellegrini2009you}: 
\begin{enumerate}[label=\alph*), noitemsep,topsep=0pt]
\item Average displacement error (ADE):  The mean square error (MSE) (Euclidean distance) over all locations of predicted trajectories and the true trajectories.
\item Average non-linear displacement error (NDE): The MSE over all locations of non-linear predicted trajectories and true trajectories.
\item Average final displacement error (FDE): The mean square error at the final predicted location and the final true location of all human trajectories.s
\end{enumerate}
\textbf{Comparison with existing methods: }We compare our results with two baselines (Linear \cite{alahi2016social,gupta2018social}, LSTM \cite{alahi2016social}) and two state-of-the-art methods (Social-LSTM [\cite{alahi2016social}, SGAN \cite{gupta2018social}):
\begin{itemize}[noitemsep,topsep=0pt]
\item Linear model Linear \cite{alahi2016social,gupta2018social} (non-LSTM) uses a linear regressor to estimate the linear parameters, minimizes the mean square error; assumes pedestrians move linearly.
\item LSTM \cite{alahi2016social} models a LSTM for each pedestrian without considering social interactions or scene information.
\item Social-LSTM \cite{alahi2016social} models the human social interactions using “social” pooling layers. We use the publicly available code given by the authors.
\item SGAN \cite{gupta2018social} models social interactions by using GAN. We use two models SGAN-20V-1 and SGAN-20VP-1, where 20V denotes that models are trained using variety loss with 20 predicted trajectories, P denotes social pooling layer. Both models generate one predicted trajectory for each pedestrian in testing phase. We used released code given in SGAN \cite{gupta2018social} to report their results.
\end{itemize}

Since the goal is to generate the best predicted trajectory, closest in  $L_2$  norm with the ground truth trajectory, we do not compare with the model SGAN-20VP-20 because this model generates 20 predicted trajectories for each pedestrian in testing phase and selects the best predicted one (the lowest ADE score compared to ground truth trajectory) which is not feasible under the problem constrains. Hence, SGAN-20VP-20 is out-of-context for our comparisons.

\noindent \textbf{Implementation details:} The implementation is done using the PyTorch framework \cite{paszke2017pytorch}. The size of all memory cells and hidden state vectors is set to 128. The network is trained with Adam optimizer \cite{kingma2014adam}, an extension to stochastic gradient descent, to update network weights during the training process. The learning rate is 0.003, and the dropout value is 0.2. The value of the global norm of gradients is clipped at 10 to ensure stable training. The model is trained on GPU Tesla P100-SXM2.

\noindent \textbf{Training. }The training is conducted in two stages:

\noindent \textit{Stage 1:} A similar “leave-one-out” approach used in \cite{alahi2016social,gupta2018social} is adopted. In details, indexing the five video sequences (ETH-Univ, UCY-Univ, UCY-Zara01, and UCY-Zara02) as $(i, j, k, l, m)$, we train (100 epochs) and validate four video sequences $(V_i, V_j, V_k, V_l)$, select the best trained model  to be used in stage 2 for the remaining (unseen) video sequence $V_m$. This process is repeated for each permutation. The data ratio for training/validation is 80/20.  

\noindent\textit{ Stage 2:} Since the scene information of each video scene is needed, the best model is further trained (in 10 epochs) on the 50\% video frames of the fifth video $V_m$. The remaining video frames are used for testing. 

\noindent \textbf{Testing:} The scene data of each grid cell in a scene and the trained network weights are fixed. We use the best trained model (weights from stage 2) and observe trajectory of each person for 8 time-steps and predict the next 12 time-steps.  

We note that the social-interaction methods only use the stage-1 training (as reported in original papers \cite{alahi2016social,gupta2018social}) to learn the social interactions. However, for reasonable comparisons we apply the same training and testing procedures for all methods. The implementation of our method will be made available. 

\subsection{Quantitative Results}
We compare our model (Scene-LSTM) with the five models described above in Table \ref{table:quantiative_results}. The results confirm that our method significantly outperforms all other methods on the three metrics: ADE, NDE, and FDE. Especially, our method predicts the final destinations (FDE) with much higher accuracy (by 0.5 meters) than the state-of-the-art SGAN-20V-1. We notice that the two models SGAN-20V-1 and SGAN-20VP-1 perform slightly better than our method in predicting non-linear trajectories (NDE) on the two video sequences ETH-Univ, UCY-Zara02. This is because our method does not capture the uncommon social interactions as they do not form common paths in these video scenes. However, the overall results validate the importance of our common human movement features in predicting future trajectories. 
\begin{table}[t]
	\caption{Quantitive results on ETH and UCY datasets (5 video sequences). All methods predict human trajectories in 12 frames using 8 observed frames. Error metrics are reported in meters (lower is better).}
	\label{table:quantiative_results}
	\centering
	\begin{tabular}{|C{1cm}||C{1.8cm}|C{1.2cm}|C{1.2cm}|C{1.5cm}|C{1.5cm}|C{1.5cm}|C{1.5cm}| }
		\hline
		Metrics & Sequences & Linear & LSTM & Social-LSTM & SGAN-20V-1 & SGAN-20VP-1  & Scene-LSTM \\
		\hline
		\multirow{6}{4em}{ADE} & ETH-Hotel  & 1.49 & 1.35 & 1.14 & 0.76 & 0.75 & \textbf{0.36} \\
		& ETH-Univ   & 2.04 & 1.97 &	2.28 & 1.26	& 1.18 & \textbf{0.95} \\ 
		& UCY-Univ   & 1.68 & 1.83 &	2.02 & 0.79	& 1.08 & \textbf{0.63} \\
		& UCY-Zara01 & 2.60 & 2.30 &	3.14 & 0.61	& 0.62 & \textbf{0.45} \\
		& UCY-Zara02	& 1.11 & 1.23 &	2.05 & 0.52 & 0.57 & \textbf{0.40} \\
		\cline{2-8} &  \textbf{Average}	& 1.78 & 1.74 &	2.13 & 0.79	& 0.84 & \textbf{0.56} \\  
		\hline 		
		\multirow{6}{4em}{NDE} & ETH-Hotel  & 3.30 & 1.71 & 2.01 & 1.66 & 1.48 & \textbf{0.76} \\ 
		& ETH-Univ   & 3.45 & 3.16 &	3.64 & \textbf{1.55} & 1.57 & 1.88 \\ 
		& UCY-Univ   & 2.22 & 2.08 &	2.36 & 1.00	& 1.21 & \textbf{0.92} \\
		& UCY-Zara01 & 2.40 & 1.75 &	2.75 & 0.71	& 0.78 & \textbf{0.65} \\
		& UCY-Zara02	& 2.67 & 2.40 &	2.87 & 0.88 & \textbf{0.81} & 0.93 \\
		\cline{2-8} &  \textbf{Average}	& 2.81 & 2.22 &	2.73 & 1.36 & 1.17 & \textbf{1.00} \\  	 
		\hline 		
		\multirow{6}{4em}{FDE} & ETH-Hotel  & 2.67 & 2.45 &	2.11 &	1.64 &	1.58 &	\textbf{0.67} \\ 
		& ETH-Univ   & 3.41 & 3.60 &	4.03 &	2.44 &	2.42 &	\textbf{1.77} \\ 
		& UCY-Univ   & 3.03 & 3.49 &	3.78 &	1.73 &	2.21 &	\textbf{1.41} \\
		& UCY-Zara01 & 4.77 & 3.98 &	5.69 &	1.32 &	1.36 &	\textbf{1.00} \\
		& UCY-Zara02	& 2.05 & 2.34 &	4.14 &	1.14 &	1.20 &	\textbf{0.90} \\
		\cline{2-8} &  \textbf{Average}	& 3.19 & 3.17 &	3.95 &	1.65 &	1.75 &	\textbf{1.15} \\  	 
		\hline 					
	\end{tabular}
	\vspace{-5mm}
\end{table}

\subsection{Ablation Study }
In this section, we present the impact of several system components (Table \ref{table:ablation}): Pedestrian Movement module using absolute locations (PM\textsubscript{abs}) vs. relative locations (PM\textsubscript{rel}), Scene Data module (SD), hard filter at the grid level (HF\textsubscript{grid}) and the subgrid level (HF\textsubscript{subgrid}), and soft filter (SF).
\begin{figure}[t]
	\centering
		\includegraphics[width=0.8\linewidth]{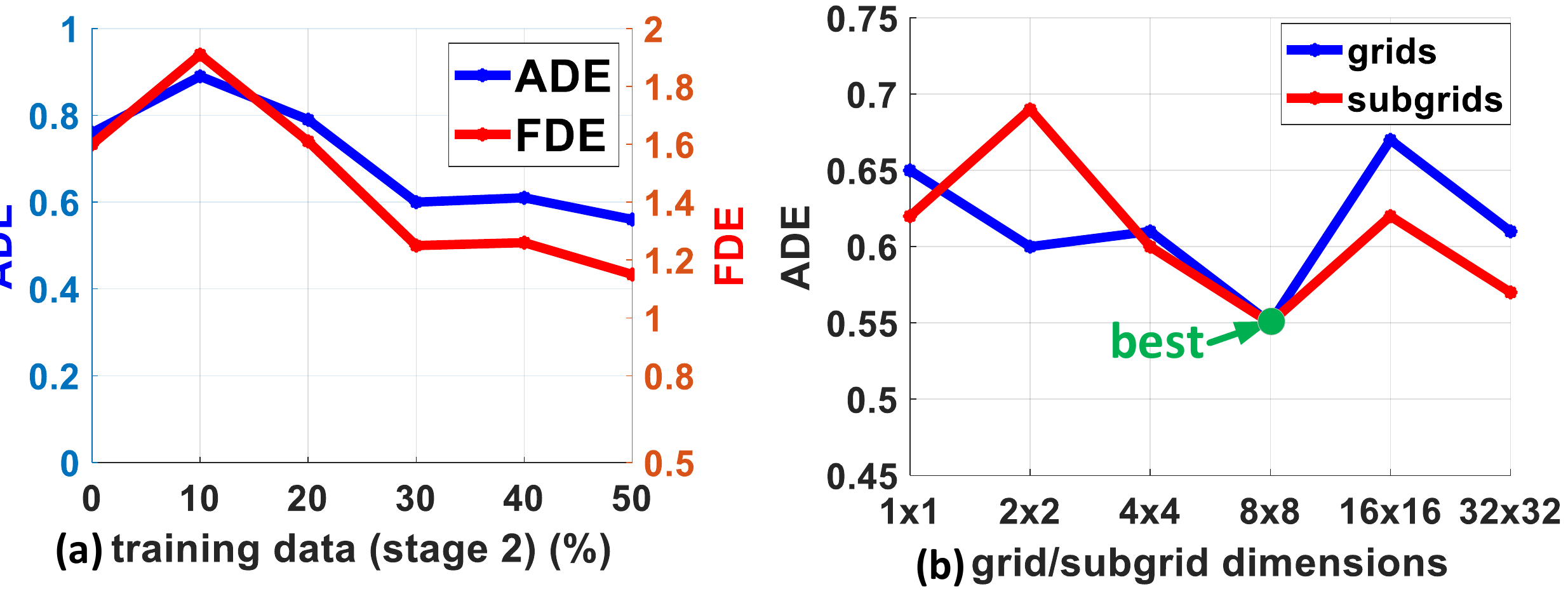}
	\caption{the impact of (a) training data amount at stage-2 and (b) finding grid/subgrid sizes to optimize prediction accuracy. The results (ADE and FDE) are calculated on average of 5 videos.}
	\label{fig:ablation}
	\vspace{-4mm}
\end{figure}

\noindent \textbf{Impact of using relative locations.} PM\textsubscript{rel} produces significantly lower errors compared to PM\textsubscript{abs}. This is because PM\textsubscript{abs} is strongly biased to a specific scene layout, while PM\textsubscript{rel} models individual target’s relative movement behavior regardless of the scene layouts; thus, PM\textsubscript{rel} allows for better transfer learning to new scenes.  

\noindent \textbf{Impact of Scene Data (SD).} As expected, using SD without the filters (the third row) worsens most of individual movement predictions.

\noindent  \textbf{Impact of hard filter at the grid cell level (HF\textsubscript{grid}).} At this level (without subgrid), the hard filter allows the Scene-LSTMs of the non-linear grid cells learn all human movements, which significantly helpful to predict the non-linear movements (lower NDE) in these cells. We also observed that the ADE is slightly increased because the scene data has a negative impact on predicting linear trajectories in the non-linear grid cells.
\begin{table}
	\caption{\textbf{Ablation study.}   PM\textsubscript{abs} and PM\textsubscript{rel}:  PM module with absolute and relative locations, respectively. SD: Scene Data module. HF\textsubscript{grid}: hard filter at the grid level.  HF\textsubscript{subgrid}:  hard filter at the subgrid level. SF: soft filter. The results (in meters) are averaged on ETH and UCY datasets (5 videos)  (lower is better).}
	
	\centering
	\begin{tabular}{l| C{0.7cm} C{0.7cm} C{0.7cm}}
		\hline
		\rowcolor{almond} Components Used  & ADE & NDE & FDE    \\
		\hline
		PM\textsubscript{abs}  			  	& 1.89 & 2.72 & 3.53 \\
		PM\textsubscript{rel} 	 		& 0.66 & 1.11 & 1.36 \\
		PM\textsubscript{rel}, SD & 0.69 & 1.34	& 1.41 \\
		PM\textsubscript{rel}, SD, HF\textsubscript{grid}		& 0.72 & 0.91 & 1.41 \\
		PM\textsubscript{rel}, SD, HF\textsubscript{grid}, HF\textsubscript{subgrid}	& 0.57 & 0.90 & 1.19 \\
		PM\textsubscript{rel}, SD, HF\textsubscript{grid}, SF & 0.62 & \textbf{0.86} & 1.30 \\ 	 			
		\hline
		PM\textsubscript{rel}, SD, HF\textsubscript{grid}, HF\textsubscript{subgrid}, SF & \textbf{0.56}	& 1.00 & \textbf{1.15} \\
		\hline
	\end{tabular}
	\label{table:ablation}
\end{table}

\noindent \textbf{Impact of hard filter at the subgrid level (HF\textsubscript{subgrid}).} The HF\textsubscript{subgrid} resolves the issue of predicting linear trajectories in HF\textsubscript{grid} and further reduces the prediction errors in all three metrics. This demonstrates the effectiveness of using the subgrid common paths as it removes uncommon paths caused by the social interactions and implicitly captures the common paths, caused by either social interactions or scene structures.

\noindent \textbf{Impact of soft filter (SF).}  Using SF at the grid cell level (the sixth row) produces more accurate non-linear trajectory predictions (lower NDE) than at the subgrid level (the last row). This is because predicting the non-linear trajectories requires more scene data obtained at grid cell level. However, considering different trajectory types (e.g. linear and non-linear) and long-term predictions, the full model (last row) still achieves the best ADE/FDE results.

\noindent \textbf{Impact of training data amount in stage 2.} To see the impact of stage 2 training data in learning common movement patterns of a new scene, an experiment is conducted by ranging the training data amounts in stage 2 from 0\% to 50\% of video frames and using the remaining 50\% data for testing (Figure \ref{fig:ablation}a). As expected, both ADE and FDE continue to decrease when more training data is used and reach the best results at 50\% video frames training.

\noindent \textbf{Impact of grid and sub-grid sizes.} The grid and subgrid sizes should be selected to best capture common human movements in each video scene. The experiment is done in two steps: (1) we first train our model on ETH and UCY datasets (5 videos) with a fixed grid size 8x8 and different subgrid sizes: 1x1, 2x2, 4x4 ,8x8, 16x16, and 32x32. The best selected subgrid size is 8x8 (\textcolor{red}{red} line, Figure \ref{fig:ablation}b) (2) We run the model again by fixing the subgrid size to 8x8 while varying grid sizes. The result, as shown in \textcolor{blue}{blue} line, confirms that the best selected grid and subgrid sizes are 8x8.  The size 8x8 indicates that the grid/subgrid size should not be too big or small; otherwise, it would not capture common human movements impacted by the scene layouts. 

\subsection{Qualitative Results}
\begin{figure}[t]
	\centering
		\includegraphics[width=\linewidth]{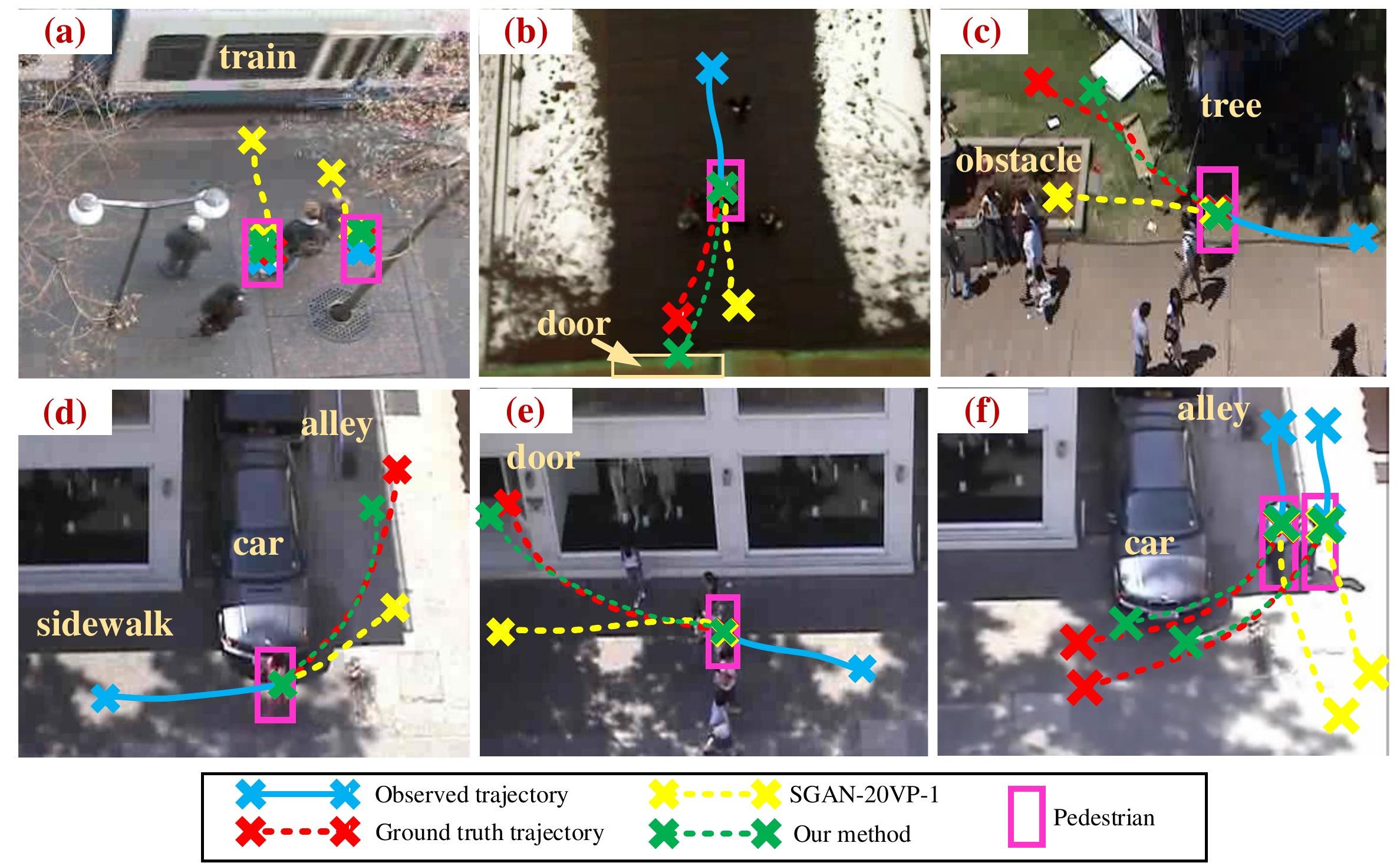}
	\caption{Qualitative comparison between our method with SGAN-20VP-1 in different scenarios: (a) pedestrians are standing still and waiting for trains, (b) a pedestrian is entering the door, (c) a pedestrian is finding a path between obstacle and trees, (d) a pedestrian makes a left-turn to the alley, (e) a pedestrian turns right to enter the building, (f) a couple turn right from the alley.}
	\label{fig:qualitative_result}
\end{figure}

 	We present qualitative comparisons with the social model SGAN-20VP-1  (Figure~\ref{fig:qualitative_result}). SGAN-20VP-1 considers the social interactions that are meaningful and socially acceptable while SGAN-20V-1 does not consider social interactions.  The visualizations show that our method generates more accurate trajectories (closer to ground truth trajectory) compared to SGAN-20VP-1 in different scene-contexts. This demonstrates the importance of learning common movements and using them to predict human movements in highly structural constrained areas. 
\begin{table}
	\caption{The quantitative results (ADE/FDE in meters) on Town Center (1 short video sequence) and PETS09S2 (3 short videos) and Grand Central (1 long video) datasets.}
	\centering
	\begin{tabular}{|C{1.1cm}|C{0.7cm}|C{1.4cm}|C{1.6cm}|C{1.4cm}| }
		\hline
		Datasets & $T_{pred}$ (s)  & SGAN-20V-1~\cite{gupta2018social}  & SGAN-20VP-1~\cite{gupta2018social}  & Scene-LSTM   \\
		\hline
		\multirow{2}{4em}{Town Center}  & 4.8 & 0.22/0.46 & 0.21/0.42 & \textbf{0.09/0.18 } \\
		& 6.4 & 0.37/0.80 & 0.38/0.81 & \textbf{0.14/0.27}  \\								
		\hline			
		\multirow{2}{4em}{PETS09 S2}      & 4.8 & 0.23/0.51 & 0.30/0.66 &\textbf{0.06/0.15}  \\
		& 6.4 & 0.43/0.93 & 0.53/1.21 & \textbf{0.11/0.23}  \\
		\hline	
		\multirow{2}{4em}{Grand Central} & 4.8 & 0.21/0.45 & 0.40/0.74 & \textbf{0.11/0.17}	\\
		& 6.4 & 0.32/0.62 & 0.79/1.50 & \textbf{0.14/0.25}  \\
		\hline		
	\end{tabular}
	\label{table:quantiative_result_3}
\end{table}
\subsection{Generalization: Evaluations on Town Center, PETS09, and Grand Central }
To present the generalizability, we further conduct experiments on new (unseen) datasets: Town Center \cite{benfold2011stable}, PETS09S2 \cite{ferryman2009pets2009}, and Grand Central \cite{zhou2012understanding}. 
\textbf{Setup.} We use the pre-trained network on ETH and UCY datasets from the previous section and further train it on 50\% of frame data of each video in this experiment (this process is similar as training stage 2 in previous experiment). The remaining frames of each video is used for testing. We generate trajectory predictions for $T_{pred} = 4.8 $ and $6.4$ seconds.
\textbf{Results.} We compare our method with two variants of SGAN \cite{gupta2018social}  as shown in Table \ref{table:quantiative_result_3}. We confirm that our method outperforms them on three datasets in ADE and FDE. For Town Center and PETS09S2 datasets, where the scenes are crowded but people mostly move linearly, the SGAN \cite{gupta2018social} method often over-predicts by considering all interactions among all pedestrians, thus fails to predict the linearity. The Grand Central dataset consists of lots of complex local social interactions, however, the common movements are paths from one train station to another, thus our method performs better by capturing these common motions. The results indicate our method can be applied to achieve the state-of-the-art results in new video sequences.

\section{Conclusion}
\label{sec:conclusion}
The novel Scene-LSTM model presented in this paper enables us to consider common human movements in localities within the scene. We have demonstrated substantial improvement in predicting trajectories using the resulting scene information, outperforming related methods. We plan to investigate fusing the scene model with social model to improve prediction quality and further explore the social interactions not only among humans but also between human and other static or moving objects. 

%
%
%
\bibliographystyle{splncs04}
\bibliography{egbib}

\end{document}